\documentclass[letterpaper, 10 pt, conference]{ieeeconf}  
\pdfoutput=1

\IEEEoverridecommandlockouts

\usepackage{flushend}
\usepackage{balance} 

\let\labelindent\relax

\usepackage{etoolbox} 
\usepackage[utf8]{inputenc} 
\usepackage[T1]{fontenc}    
\usepackage[english]{babel} 
\usepackage[protrusion=true,expansion=true]{microtype}
\usepackage{comment}
\usepackage{cancel}
\usepackage{csquotes}
\usepackage{listings}
\usepackage{nameref}
\usepackage{paralist}
\usepackage{xspace}
\usepackage{setspace}
\usepackage{makeidx}
\usepackage{pdfpages}

\usepackage{amsmath,amssymb} 
\usepackage{amsfonts}       
\usepackage{mathtools}
\usepackage{array} 
\usepackage{nicefrac}       
\usepackage{breqn}          
\usepackage{textcomp}
\usepackage{bm} 
\usepackage{pifont}
\usepackage[
  separate-uncertainty = true,
  multi-part-units = repeat
]{siunitx}

\usepackage{graphicx}
\usepackage{adjustbox} 
\usepackage{epsfig}

\usepackage{float}
\usepackage{subcaption}
\usepackage[font=small,labelfont=bf]{caption}
\usepackage{lscape}      

\usepackage{tikz}  
\usepackage{makecell}
\usepackage{overpic}
\usepackage{algorithm}
\usepackage[noend]{algpseudocode}

\usepackage{array} 
\usepackage{tabularx}
\usepackage{multirow}
\usepackage{multicol}
\usepackage{booktabs}   

\usepackage{paralist}
\usepackage{enumitem}
\setlist[itemize]{noitemsep,leftmargin=*,topsep=0in}
\setlist[enumerate]{noitemsep,leftmargin=*,topsep=0in}

\usepackage{url}            
\PassOptionsToPackage{table}{xcolor}
\usepackage{color,colortbl} 
\usepackage{soul}

\PassOptionsToPackage{capitalize, noabbrev}{cleveref}
\usepackage[pagebackref=false,breaklinks=true,colorlinks,urlcolor=blue,citecolor=blue,linkcolor=blue,bookmarks=false]{hyperref}

\makeatletter
\let\NAT@parse\undefined
\makeatother
\usepackage[numbers,sort&compress]{natbib}

\usepackage{blindtext}
\usepackage{bbm}

\usepackage{lipsum}

\usepackage{titlesec}
\titlespacing{\section}{0pt}{0.4\baselineskip}{0.25\baselineskip}
\titlespacing{\subsection}{0pt}{0.25\baselineskip}{0.15\baselineskip}
\titlespacing{\subsubsection}{0pt}{0.05\baselineskip}{0.03\baselineskip}

\renewcommand{\paragraph}[1]{\vspace{1em}\noindent\textit{#1} --}

\newcommand{\algoName}{ActAIM\xspace}

\title{\LARGE \bf
Self-Supervised Learning of Action Affordances as Interaction Modes
}

\author{
    Liquan Wang$^{\dagger}$, %
    Nikita Dvornik$^{\ast}$, %
    Rafael Dubeau$^{\dagger}$, %
    Mayank Mittal$^{\ddagger\star}$, %
    Animesh Garg$^{\dagger\star}$ %
    \thanks{
    Correspondence to: \texttt{garg@cs.toronto.edu}}
    \thanks{
    $^{\dagger}$University of Toronto \& Vector Institute,  
    $^{\ddagger}$ETH Zurich, 
    $^{\star}$Nvidia,
    $^{\ast}$Samsung.}
}

\usepackage{color,soul}

\definecolor{chartblue}{RGB}{21, 53, 98}

\begin{document}

\maketitle
\thispagestyle{empty}
\pagestyle{empty}

\begin{abstract}
When humans perform a task with an articulated object, they interact with the object only in a handful of ways, while the space of all possible interactions is nearly endless.
This is because humans have prior knowledge about what interactions are likely to be successful, i.e., to open a new door we first try the handle.
While learning such priors without supervision is easy for humans, it is notoriously hard for machines.
In this work, we tackle unsupervised learning of priors of useful interactions with articulated objects, which we call \emph{interaction modes}.
In contrast to the prior art, we use no supervision or privileged information; we only assume access to the depth sensor in the simulator to learn the interaction modes.
More precisely, we define a successful interaction as the one changing the visual environment substantially and learn a generative model of such interactions, that can be conditioned on the desired goal state of the object.
In our experiments, we show that our model covers most of the human interaction modes, outperforms existing state-of-the-art methods for affordance learning, and can generalize to objects never seen during training.
Additionally, we show promising results in the goal-conditional setup, where our model can be quickly fine-tuned to perform a given task.
We show in the experiments that such affordance learning predicts interaction which covers most modes of interaction for the querying articulated object and can be fine-tuned to a goal-conditional model. For supplementary: \url{https://actaim.github.io/}.
\end{abstract}
\section{Introduction}

Humans demonstrate tremendous flexibility in operating objects around them.
By leveraging \textit{prior} experiences, we can adapt and manipulate new objects through careful interactions or exploration.
A standard method in robotics for building object priors is by hand-crafting models based on our knowledge of an object's relevant properties for a given task.
For instance, various works in articulated object manipulation design modules to detect handles, obtain part-wise object segmentation, and estimate articulation parameters to define interaction plans~\cite{abbatematteo2020kinematicmodels,klingbeil2010open,mittal2021articulated}.
The rigidness in these explicit models limits their ability to generalize and capture novel ways of changing the state of an unknown object, such as grasping its edge or pushing its surface.
In this work, we aim to build representations that allow defining these \emph{interaction modes} of an object implicitly, thereby providing \textit{prior} knowledge for manipulating unseen objects.
To do so, we focus on interacting with articulated objects with multiple moving parts as they do provide multiple affordances to be discovered.

\begin{figure}
    \centering
    \includegraphics[width=\linewidth]{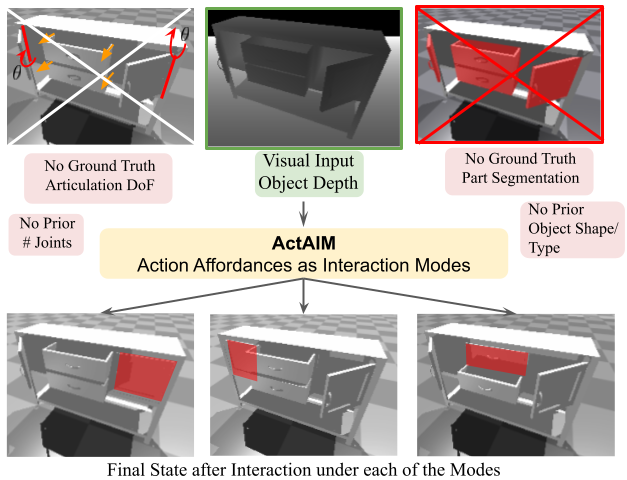}
    \caption{
    \textbf{ActAIM Overview.}
    We propose a novel model architecture to learn action affordances of articulated objects using purely visual data. Without the use of any privileged ground-truth information or explicit supervision, our model learns distinct interaction modes that can generalize to new unseen objects.
    }
    \label{fig:my_label}
\end{figure}

Discovering \emph{interaction modes} of an object has often been connected to the idea of \textbf{affordances} \cite{gibson:hal-00692033,gibson1986ecological}, what does the object offer to an actor in the environment.
For instance, a cabinet with three drawers has six possible interaction modes; each drawer providing the affordance to open or close it. 
Supervised learning approaches require collecting a large dataset of interactions, typically with a random policy, and labeling their effect as a change in the object's state~\cite{mo2021where2act,wu2021vat,do2018}. 
However, the sparsity of events that cause a noticeable change leads to a huge data imbalance and makes these approaches sample inefficient.
Reinforcement learning (RL) methods avoid large-scale data annotation but suffer from a similar exploration issue. The policy tends to exploit only a limited region of the object for manipulation, thereby failing to discover all the interaction modes~\cite{xu2022umpnet}.

Besides the exploration issue in discovering interaction modes, existing works \cite{KessensRSBG10,opendoor,servicerobots,generalized,kinematicefficient,mittal2021articulated,screwnet,kinematicmodels,learnopendoor} use the object's state information directly as part of their observations, rewards function computation, or for scoring the amount of change caused by a particular action.
However, humans primarily learn and act in partially observed settings. Relying solely on visual information exacerbates the learning problem since discriminating between interaction modes from images only (not using additional privileged information) is challenging. 

In this paper, we present \algoName (\textbf{Act}ion \textbf{A}ffordances as \textbf{I}nteraction \textbf{M}odes) to overcome these issues by introducing the concept of interaction modes which can be clustered with the specific feature encoder and using only the visual observations during training. 
\algoName discovers semantically meaningful \& varied interaction modes and is also able to provide goal-conditional task completion.
We use implicit geometry feature to build the semantic representation of the object instance which helps generalized across different categories.
Our key contributions are as follows:
\begin{enumerate}
  \item We propose the idea of interaction modes and a method to learn meaningful interactions in a self-supervised manner, from visual observations only. 
  \item We propose a new clustering-based data collecting strategy that increases the diversity among interaction modes in the collected data.
    \item We experimentally show that \algoName generates actions that cover a variety of ground-truth interaction modes and lead to successful goal completion, when conditioned on the goal observation (in the goal-conditional setup).
\end{enumerate}
\section{Related Work}

Affordance is an important concept in many fields. 
Based on the definition from Gibson \cite{gibson:hal-00692033,gibson1986ecological}, there has been a considerable amount of research on affordance in psychology, neuroscience, cognitive science, human-computer interaction, robotics, vision, and design. 
Researchers captured this general notion in various ways, including language, semantic segmentation, and key points.  

\paragraph{\textbf{Affordance in robot learning}} 
Prior works have shown that learning to solve the manipulation problem could benefit from understanding the concepts of affordance. 
\cite{affordancecue, affordancegeometric, affordanceinfer, affordancegrasp} focus on extracting affordance features using neural networks directly from image observation in supervised learning.
Furthermore, semantic segmentation could be further extended to scenes with multi-object in ~\cite{affordanceseg}.
Besides, affordance can be also defined as the probability of transition function representing the possibility of taking action in a certain area in ~\cite{whatcanido}. 
~\cite{hierarchical2019manoury} defines affordance with primitive actions and trains the agent to learn feasible action in different states which boosts the efficiency and scalability in performance. 
Inspired by the above papers, our work learns affordance from visual input by defining proper action primitives and trains the model without any supervision or privileged information from humans.

\paragraph{\textbf{Semantic keypoint discription}} 
An explicit affordance representation is the keypoint representation
\cite{manuelli2019kpam,turpin2021gift,qin2019keto,fang2018learning}. 
Using the two-gripper robot, with the help of existing grasping algorithms such as \cite{sundermeyer2021contactgraspnet,mahler2017dexnet}, keypoint representation can be used to guide robotic grasping in tool or articulated object manipulation tasks. 
For task-directed affordance, keypoint representation is sufficient \cite{qin2019keto,mahler2017dexnet,turpin2021gift} since the choice of interaction point is required to be limited and robotic moving trajectory is compatible to decompose into point movement. 
Considering the multi-mode interaction discovery task, we picked a certain keypoint as the interaction point conditional on the interaction modes combining with the dense local geometry feature following the ideas from \cite{khetarpal2021temporally,khetarpal2020i,danfei2021}. 

\paragraph{\textbf{Dense pixel-wise affordance}} 
Existing dense pixel-wise affordance learning papers such as \cite{mo2021where2act,myers2015,do2018,mo2021o2oafford} predict per-pixel affordance labels and query from these encodings. 
Among these papers, \cite{mo2021where2act} and \cite{jiang2021synergies} enlightened us to combine implicit representation in articulated object manipulation. 
\cite{jiang2021synergies} solves the grasping task using the Convolutional Occupancy Network \cite{peng2020convolutional} implicit representation model and explores the synergy of using geometric features.
But \cite{jiang2021synergies} is limited to grasping tasks and relies on privileged information such as object meshes. 
\cite{mo2021where2act} uses pure visual input to predict interaction points on articulated objects with predefined action primitive but requires fixed modes and part segmentation.
Our work takes advantage of \cite{jiang2021synergies} by using the neural implicit representation to extract local geometry features to help generalize among different articulated objects.

\section{Problem Formulation}
\label{sec:problem}

Inspired by the concept of prior knowledge in manipulation, we formulate the problem of discovery of object affordance in an unsupervised setting without access to privileged information. The goal is to build object-centric priors using only perception throughout the learning process, such that they facilitate: 1) realizing different types of \emph{interation modes} an object offers implicitly, and 2) capturing \emph{where} and \emph{how} to interact with an object for a given interaction mode.

ActAIM takes the depth image of an object as input and outputs the possible actions that can be executed with the object through interaction modes.  
In contrast to Where2Act~\cite{mo2021where2act} that discretizes the action space into six primitives, we consider a continuous action space for a parallel-jaw robotic gripper, $a = (\mathbf{p}, \mathbf{R}, \mathbf{F})$. The primitive action first reaches and attempts to grasp an interaction point $\mathbf{p}$ over the visible articulated parts of the object $P$ with an orientation $\mathbf{R} \in SO(3)$, and then moves a certain fixed distance along a unit direction $\mathbf{F} \in [-1, 1]^3$.

Formally, we obtain a prior distribution over possible interactions with an articulated object leading to a change in its state. 
Under a partially observed setting, this distribution can be denoted as $\mathbb{P}(a | o)$, where $o$ is a visual observation of the articulated object, such as its depth image $D$, point cloud $P$ or truncated signed distance field (TSDF) representation $V$.
%
To discriminate between different interaction modes for an articulated object, we introduce a latent variable $z \in Z$, and write the prior as:
\begin{align}
  \mathbb{P}(a|o) = \underbrace{\mathbb{P}(a|o,z)}_{\text{action predictor}}~ \underbrace{\mathbb{P}(z|o)}_{\text{mode selector}}
  \label{eq:prob_actions}
\end{align}

The distribution $\mathbb{P}(z|o)$ models the possible interaction modes (latent affordance) of an object given its current observation. For instance, a completely closed cabinet can only be opened, so only half of the maximum interaction modes associated with it are feasible.
The distribution $\mathbb{P}(a|o,z)$ models the actions that would lead to the change associated with the interaction mode $z$.


\section{\algoName : Learning Interaction Modes}

Our goal is to obtain the distributions in \eqref{eq:prob_actions} without requiring explicit supervision labels or reward signals, which are typically computed using articulated objects' joint state. To this end, we propose a self-supervised learning method that generates its own labels through interacting with an object and uses these to learn a common visual embedding for articulated objects that generalizes over unseen articulated object categories and instances.

\begin{figure}[t]
    \centering
    \includegraphics[width=\linewidth]{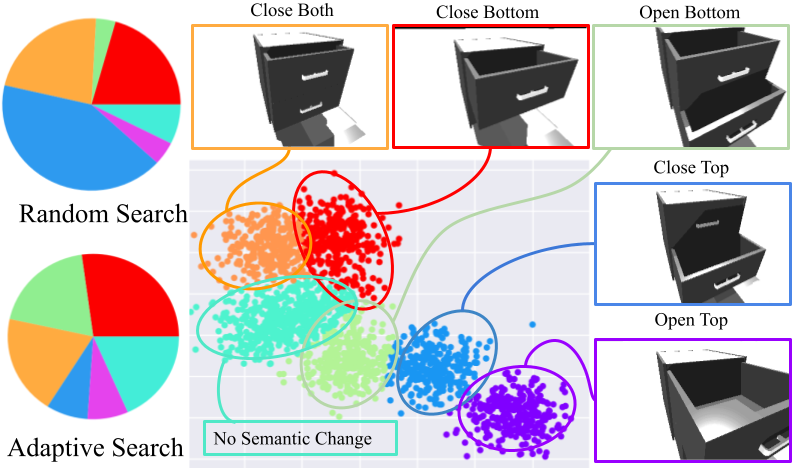}
    \caption{
    \textbf{Adaptive data collection using GMM.}
    Visualization of using GMM to cluster different interaction modes. The clustering is presented with the t-SNE projection of the manifold. It illustrates that different interaction modes can be distinguished and clustered using simple image encoding. The two pie charts on the left show that using adaptive search for data collection increases the proportion of rare modes of interaction.
    }
    \label{fig:data-collection}
\end{figure}

\subsection{Adaptive data collection}
\label{subsec:data}

We start the data collection by executing actions from the random policy 
in the simulator.
For every executed action primitive, we store the tuple $(o, a, o', \hat{y})$, where $o = (D_0, V)$ is the depth image and TSDF at the initial state of the articulated object, $a$ is the executed action primitive parameters, $o'=(D_1)$ is the depth image of the articulated object after the interaction, and $\hat{y}$ is the computed label determining whether the interaction is successful. We compute the TSDF before interaction using multi-view depth images~\cite{zeng20163dmatch}.


To collect diverse interaction data, we need to vary  object categories, instances, their initial states, camera views, and action primitive parameters $a$.
Sampling actions with the random policy is sub-optimal due to the poor coverage of all  possible interaction modes (e.g., it's harder to randomly pull the handle than to push the door). Hence, we propose an adaptive scheme using unsupervised learning to improve data more balanced across different modes of interaction.

First, to get an embedding function for the depth images, we train an autoencoder, $\hat{D}=\mathcal{D}_D(\mathcal{E}_D(D))$, using an $L_2$ reconstruction loss.
In the rest of the paper, we use the encoder representations, i.e., $\mathcal{E}_D(D)$, not the raw depth maps $D$, as subsequent inputs.
Then, we fit a multi-modal distribution that clusters different interactions into, presumably, interaction modes.
For this, we use the Gaussian Mixture Models (GMMs): $\mathbb{P}(a|D_0, \theta) = \sum^{K}_{k=1}\alpha_k p(a|D_0,\theta_k)$ where $\theta$ is the distribution parameters and $K$ is the maximum number of mixtures. 
We fit this GMM iteratively for a single object instance with a fixed initial state.
We define the effect of an executed action as, $\tilde{\tau} = \mathcal{E}_D(D_1) - \mathcal{E}_D(D_0)$. At the start, we collect interaction data of size $M$ using a random policy.
We fit GMM only on data that leads to a change in the embedded space: $\{(D_0, a, D_1), | ||\tilde{\tau}||_2 \geq \lambda \}$, where $\lambda$ is a fixed threshold. 
We determine $\hat{y} = 1$ if $||\tilde{\tau}||_2 \geq \lambda $ otherwise $\hat{y} = 0$.
For the following iterations, we sample $\epsilon M$ interactions with a random policy and $(1-\epsilon) M$ interactions from the GMM, and fit a GMM again over the collected data.
As shown in Fig.2, clustering using GMM discriminates among different interaction modes and this adaptive sampling strategy yields a more balanced coverage of interaction modes compared to a random policy. 
\begin{figure}[!t]
    \centering
    \includegraphics[width=\linewidth]{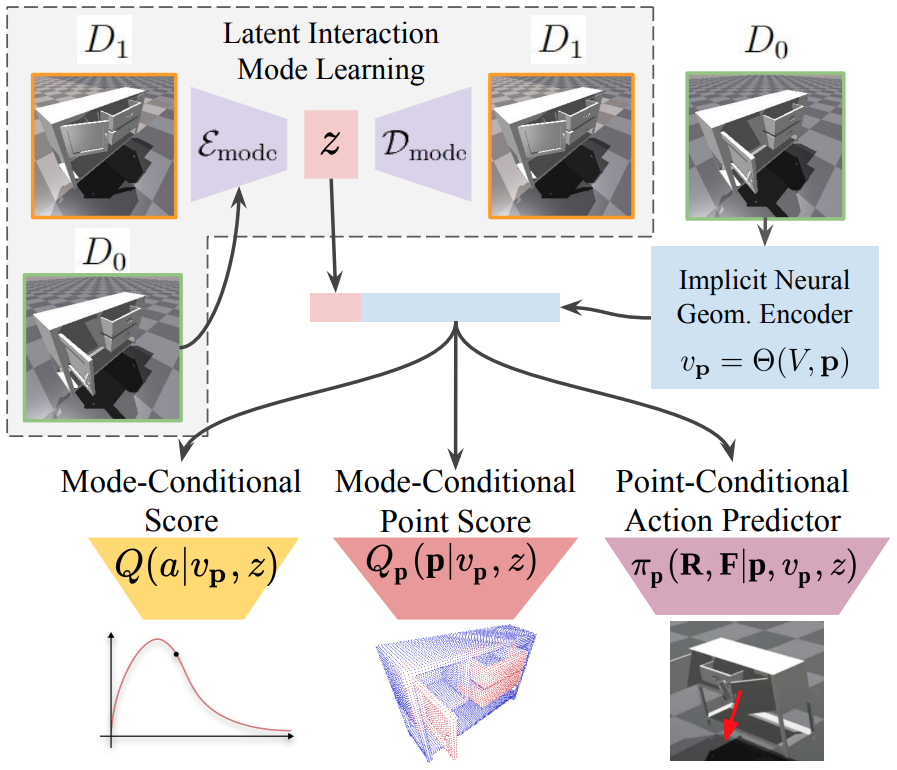}
    \caption{ \textbf{\algoName training overview.}
    This figure shows the process of model training. During the training, \algoName takes in the initial and final depth image observations $D_0$ and $D_1$. $D_0$ and $D_1$ are passed into the encoder $\mathcal{E}_{\text{mode}}$ to produce the mode latent $z$, and $D_0$ is passed into the implicit neural geometry encoder to produce a local geometry feature $v_{\mathbf{p}}$. Mode latent $z$ and local feature $v_{\mathbf{p}}$ are passed into the mode-conditional score function, mode-conditional point score, and point-conditional action predictor to predict scoring over the point cloud and the action $\mathbf{R}$ and $\mathbf{F}$.
    }
    \label{fig:2-pipeline}
\end{figure}

\subsection{Self-supervised model for interaction discovery}
\label{subsec:model}

While the GMM approach described for data generation clusters different interaction modes, it only does so for single object instance at a specific state.
Instead, we want a model that generalizes across poses, instances and categories of objects.
Given that $a = (\mathbf{p}, \mathbf{R}, \mathbf{F})$, we can redefine~\eqref{eq:prob_actions} as:
\begin{align}
    \mathbb{P}(a|o,z) = %
    \mathbb{P}_{\mathbf{R}, \mathbf{F}|\mathbf{p}}(\mathbf{R}, \mathbf{F}|\mathbf{p}, o, z)~ 
    \mathbb{P}_{\mathbf{p}} (\mathbf{p}|o,z)
\end{align}
where $\mathbb{P}_{\mathbf{p}}$ defines the probability of selecting interaction point $\mathbf{p}$ under an interaction mode $z$, and $\mathbb{P}_{\mathbf{R}, \mathbf{F}|\mathbf{p}}$ defines that for the gripper rotation and moving direction given a selected interaction point $\mathbf{p}$ and interaction mode $z$.
This decomposition of the action predictor helps reducing the sampling complexity during inference.
Our implementation of this decomposed pipeline is given in Fig.~\ref{fig:2-pipeline}.

\begin{figure*}
    \centering
    \begin{minipage}{0.68\linewidth}
    \includegraphics[width=\linewidth]{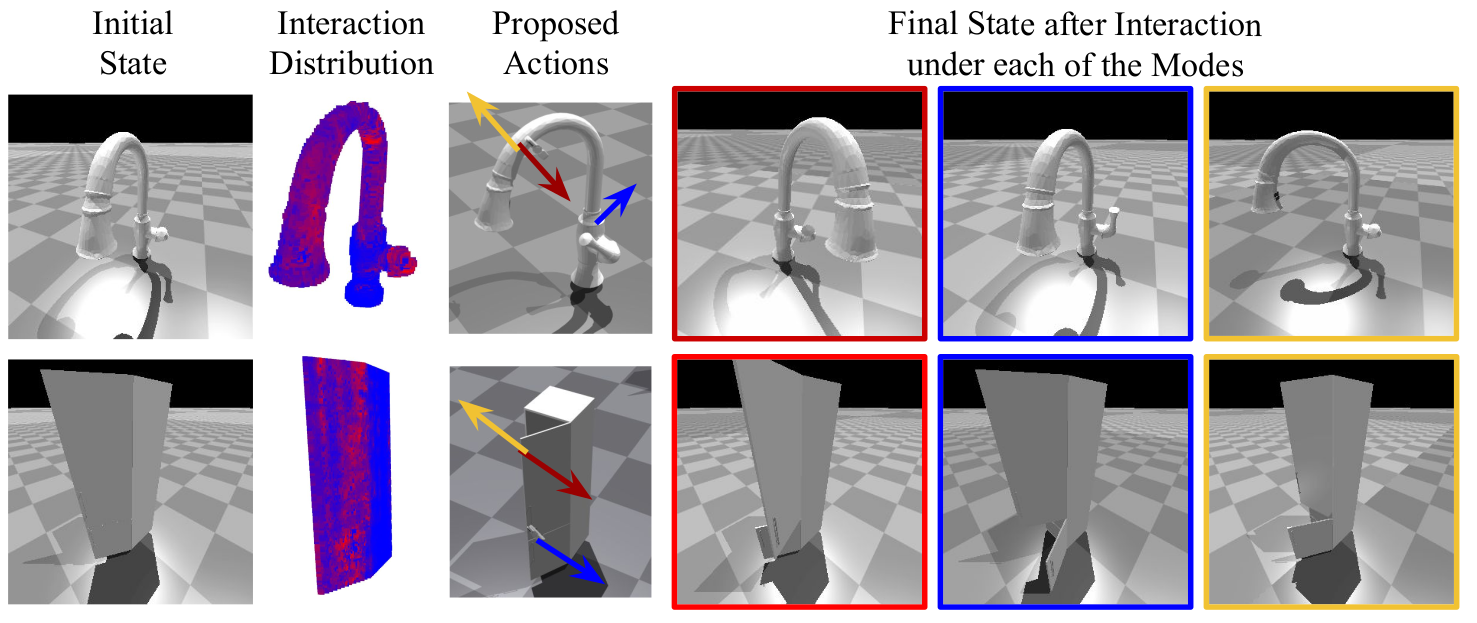}
    \end{minipage}
    \begin{minipage}{0.30\linewidth}
    \caption{\textbf{Generative model object manipulation results: } \algoName takes in the initial state and predicts the interaction distribution based on the interaction point scoring model. The distribution shows the valid part (colored in red) to interact with. We sample interaction point based on the point score and execute the predicted action. The action will lead to different interaction mode as shown.}
    \label{fig:self-supervised}
    \end{minipage}
\end{figure*}

\subsubsection{Mode selector $\mathbb{P}(z|o)$}
We model the mode selector $\mathbb{P}(z|o)$ with a Conditional Variational Autoencoder (CVAE) \cite{cvae} with encoder $\mathcal{E}_{\text{mode}}$, decoder $\mathcal{D}_{\text{mode}}$.
This autoencoder operates on the embeddings of the depth images, $(\mathcal{E}_D(D_0), \mathcal{E}_D(D_1))$, as described above, to generate the latent interaction mode $z$. 
Latent interaction mode $z$ captures the difference between $D_0$ and $D_1$ which represents the state change. 
The decoder $\mathcal{D}_{\text{mode}}$ takes in the conditional variable $\mathcal{E}_D(D_0)$ and the latent interaction mode $z$ to reconstruct $\mathcal{E}_D(D_1)$.
We train this CVAE structure together with the further model and optimize it with the regularization loss and reconstruction loss.

\subsubsection{Implicit Neural Geometry Encoder}
We use an implicit neural geometry encoder that encodes local geometry features to improve the generalizability of the model over different categories of articulated objects.
Implicit representation is a continuous function with neural network as the input. 
We formalize local geometry feature extraction as $v_{\mathbf{p}} = \Theta(V, \mathbf{p})$ where $V$ is the TSDF of the object and $\mathbf{p}$ is a queried point. 
The structure of the feature extraction network, $\Theta$, is adapted from Convolutional Occupancy network \cite{peng2020convolutional} (ConvOnet).
The ConvOnet decoder conveys the core idea of implicit representation which provide memory-sufficient storage of point-wised feature data.
Consider the plane feature $\Omega = \{\Omega_{xy}, \Omega_{yz}, \Omega_{xz}\}$ from the ConvOnet, 
we extract the point specific feature $v_{\mathbf{p}}$ given the querying point $\mathbf{p}$. 
Following \cite{jiang2021synergies}, we concatenate features using querying point projected on the corresponding plane and perform bilinear interpolation $\phi$ around the neighborhood of the querying point $\mathbf{p}$. 
We represent our local feature extraction as, 
\begin{align}
  v_{\mathbf{p}} = [\phi(\Omega_{xy}, \mathbf{p}_{xy}), \phi(\Omega_{yz}, \mathbf{p}_{yz}), \phi(\Omega_{zx}, \mathbf{p}_{zx}))]
\end{align}
Where $\mathbf{p}_{xy}$ denotes the x and y coordinate of the point $\mathbf{p}$.
We combined the action predictor with this Implicit Neural Geometry Encoding and express the action predictor as, 
\begin{align}
    \mathbb{P}(a|o,z) & = %
    \mathbb{P}_{\mathbf{R}, \mathbf{F}|\mathbf{p}}(\mathbf{R}, \mathbf{F}|\mathbf{p}, v_{\mathbf{p}}, z)~
    \mathbb{P}_{\mathbf{p}} (\mathbf{p}, v_{\mathbf{p}} | o,z)
\end{align}


\subsection{Training procedure}
\label{subsec:training}

Through the above formulation, we want to jointly train the distributions ($\mathbb{P}_{\mathbf{p}}$, $\mathbb{P}_{\mathbf{R,F} | \mathbf{p}}$), the CVAE ($\mathcal{E}_{\text{mode}}, \mathcal{D}_{\text{mode}}$), and the neural geometry encoder ($\Theta$). 

\paragraph{Mode-conditional score function $Q(a|o, z)$} 
We regard the distribution $\mathbb{P}(a|o,z)$ as a score function $y = Q(a|v_{\mathbf{p}}, z)$ to denote the probability of success when taking action $a$ with the local geometry feature $v_{\mathbf{p}}$. 
We use the data collected from GMM represented as $\{D_0, D_1, a, \hat{y}\}_i$ and $z$ from the mode selector CVAE model. 
We optimize this module using the binary cross entropy loss with self-supervised label $\hat{y}$ from the dataset. 
The predicted mode conditional critic will be used to help evaluate the point score function later.

\paragraph{Mode-conditional point score function $Q_{\mathbf{p} }(\mathbf{p} | o, z)$} We model the distribution $\mathbb{P}_{\mathbf{p}}$ (of successful interaction at point $p$) as a score function $y_{\mathbf{p}}=Q_{\mathbf{p}}(\mathbf{p} | v_{\mathbf{p}}, z)$, taking the local geometric feature $v_{\mathbf{p}}$ as input.
We train $Q_{\mathbf{p}}$ using the data $\{D_0, a\}_i$ and the computed $z$, and optimize with binary cross entropy loss. 
The ground-truth $\hat{y}_{\mathbf{p}}$ referring the probability of point $\mathbf{p}$ becoming the interaction point. 
To obtained the training signal $\hat{y}_{\mathbf{p}}$, we randomly sample N (100) rotations $\mathbf{R}_i$ and moving directions $\mathbf{F}_i$ and compute the label with mode conditional score function $Q(o,a|z)$ as followed, 
\begin{align}
  \hat{y_{\mathbf{p}}} = \text{max} \{Q(\hat{\mathbf{R}}_i, \hat{\mathbf{F}}_i, \mathbf{p}|o, z)| i = 1,...,N\}
\end{align}
We are using the maximum to express that the point is valid to interact with once there exists a successful interaction using this interaction point.

\paragraph{Point-conditional action predictor $\pi_{\mathbf{p}}(\mathbf{R}, \mathbf{F}|\mathbf{p}, o, z)$}
Given the interaction point $\mathbf{p}$ sampled from the $Q_{\mathbf{p}}$, \algoName predicts the rotation and moving direction with Point conditional action predictor $\pi_{\mathbf{p}}(\mathbf{R}, \mathbf{F}|\mathbf{p}, v_{\mathbf{p}}, z)$ together with the local implicit geometry feature.
The module produces rotation $\mathbf{R}$ and moving direction $\mathbf{F}$ directly and optimized with collected data  $\{D_0, D_1, a\}_i$. 
Denoting the ground-truth rotation and moving direction as $\hat{\mathbf{R}}$ and $\hat{\mathbf{F}}$, the loss can be written as
\begin{align}
  \mathcal{L}_{\mathbf{R}} + \mathcal{L}_{\mathbf{F}}
  =(\mathbf{F} - \hat{\mathbf{F}})^2
  + ( 1 - |\mathbf{R} \cdot \hat{\mathbf{R}}|)
\end{align}

\paragraph{Final Loss}
The complete training loss is now denoted as
\begin{align}
    \mathcal{L} = \mathcal{L}_{\text{CVAE}} + \mathcal{L}_{Q} + \mathcal{L}_{Q_{\mathbf{p}}} + \mathcal{L}_{\mathbf{R}} + \mathcal{L}_{\mathbf{F}}
\end{align}

\begin{table*}[t] \LARGE
  \centering
  \caption{\small
  \textbf{Self-Supervised Affordance Mode Discovery:}
  We evaluate our design choices with baseline and ablation study using the metrics of sample-success rate, weighted modes ratio and normalized conditional entropy. We also illustrated an extra column which is the average of the section of articulated objects. In each column section, we bold the best numbers and show that our model outperforms most of the time.
  {\footnotesize Categories: 
  \hl{faucet, table, cabinet, door,  window, fridge},
  \hl{kitchen pot, kettle, switch}
  }
  }
  \label{table:self-supervised}
  \resizebox{1.0\linewidth}{!}
  {
  \begin{tabular}{l||ccccccc|ccccccc|cccc}
  \toprule
\rowcolor[HTML]{FFDEB4}
\textbf{Test Set}
& \multicolumn{7}{c|}{\textbf{Unseen States of Training Objects}} & \multicolumn{7}{c|}{\textbf{Unseen instances of Training Categories}} & \multicolumn{4}{c}{\textbf{Unseen Categories}}\\ 
  \midrule

\cellcolor[HTML]{ACCEAA}    \textbf{Sample Success Rate \% $\uparrow$} & 
\cellcolor[HTML]{CBCEAA}    \includegraphics[width = 0.042\linewidth]{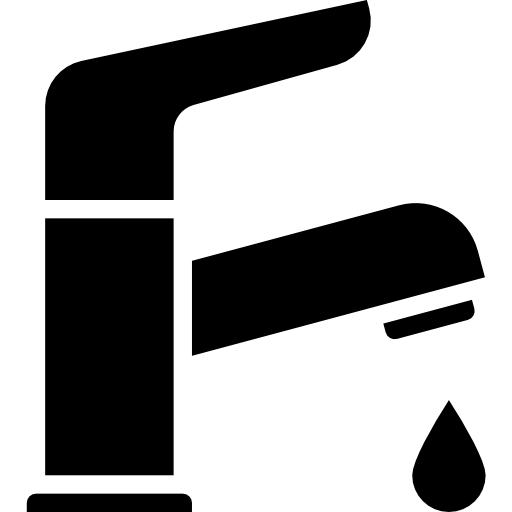} &
\cellcolor[HTML]{CBCEAA}    \includegraphics[width = 0.042\linewidth]{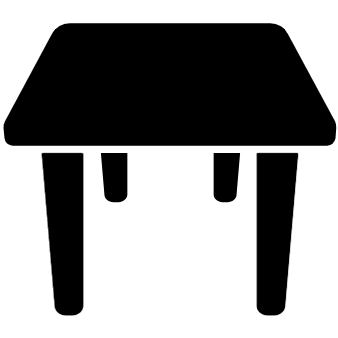} &
\cellcolor[HTML]{CBCEAA}    \includegraphics[width = 0.042\linewidth]{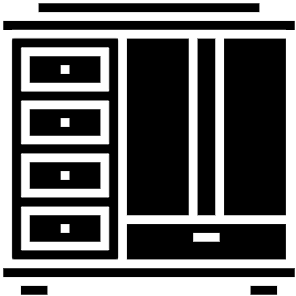} &
\cellcolor[HTML]{CBCEAA}    \includegraphics[width = 0.042\linewidth]{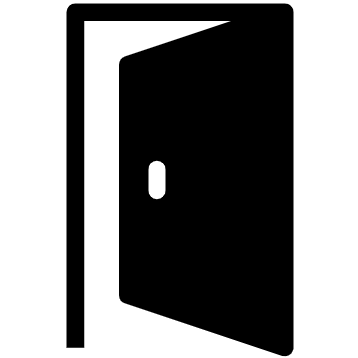} &
\cellcolor[HTML]{CBCEAA}    \includegraphics[width = 0.042\linewidth]{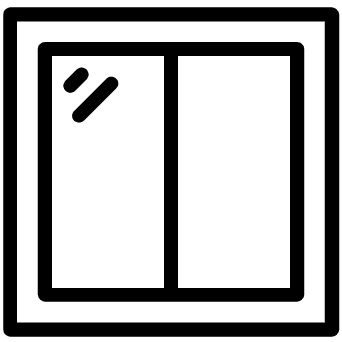} &
\cellcolor[HTML]{CBCEAA}    \includegraphics[width = 0.042\linewidth]{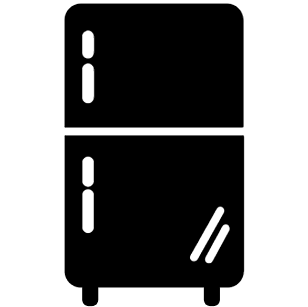} &
\cellcolor[HTML]{ABCEBF}    \includegraphics[width = 0.042\linewidth]{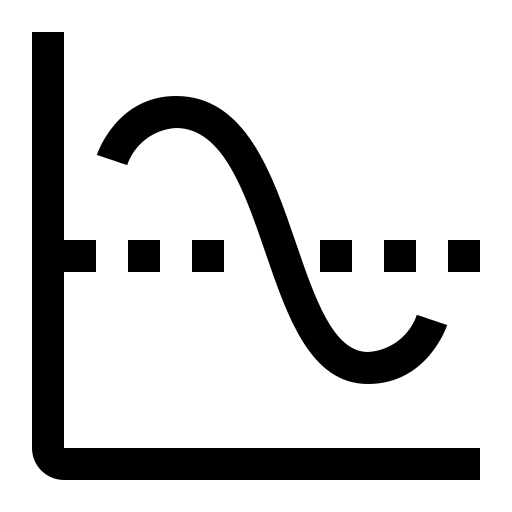} &

\cellcolor[HTML]{CBCECC}    \includegraphics[width = 0.042\linewidth]{faucet.png} &
\cellcolor[HTML]{CBCECC}    \includegraphics[width = 0.042\linewidth]{noun_Table_59987.png} &
\cellcolor[HTML]{CBCECC}    \includegraphics[width = 0.042\linewidth]{noun_Cabinet_2881254.png} &
\cellcolor[HTML]{CBCECC}    \includegraphics[width = 0.042\linewidth]{noun_Door_1549119.png} &
\cellcolor[HTML]{CBCECC}    \includegraphics[width = 0.042\linewidth]{noun_window_3203560.png} &
\cellcolor[HTML]{CBCECC}    \includegraphics[width = 0.042\linewidth]{noun_Fridge_1875643.png} &
\cellcolor[HTML]{ABCEBF}    \includegraphics[width = 0.042\linewidth]{avg.png} &
    
\cellcolor[HTML]{CBCEEE}    \includegraphics[width = 0.042\linewidth]{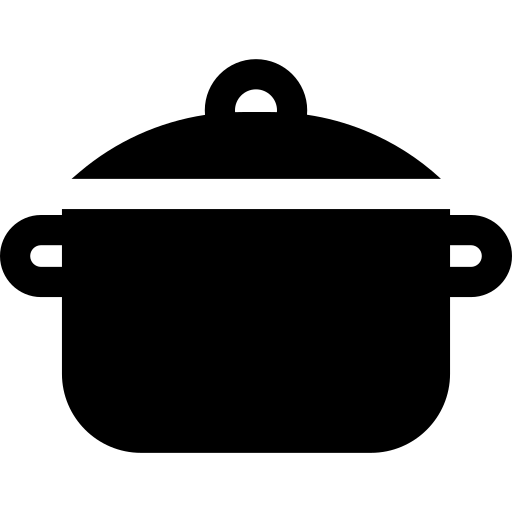} &
\cellcolor[HTML]{CBCEEE}    \includegraphics[width = 0.042\linewidth]{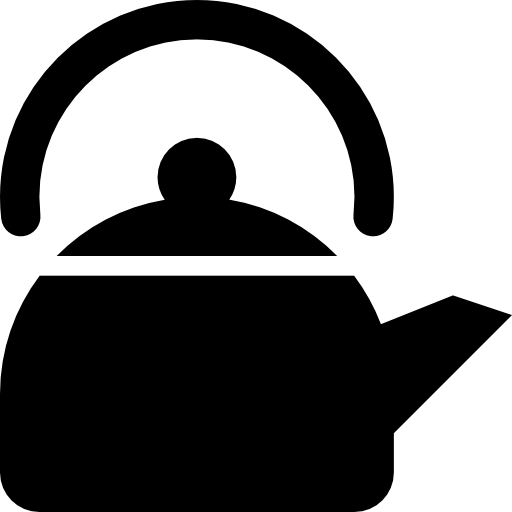} &
\cellcolor[HTML]{CBCEEE}    \includegraphics[width = 0.042\linewidth]{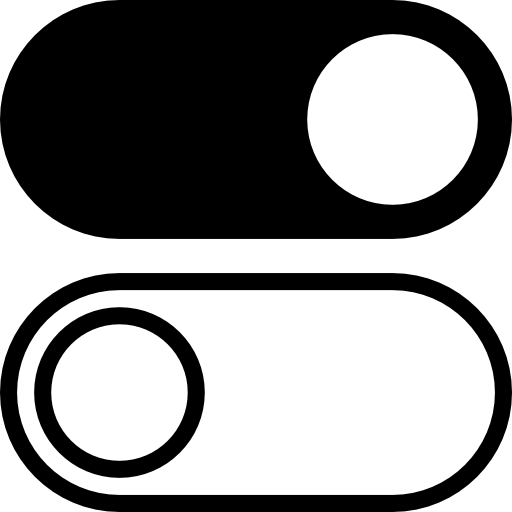} & 
\cellcolor[HTML]{ABCEBF}    \includegraphics[width = 0.042\linewidth]{avg.png} \\
  
\midrule
  Random Interaction & 5.61 &6.05 &8.47 &14.59 &18.15 &5.66 &9.76 &7.99 &3.38 &6.47 &13.86 &20.92 &4.51 &9.52 &13.32 &10.42 &6.04 &9.93  \\
  \rowcolor[HTML]{EFEFEF} 
  Where2Act \cite{mo2021where2act} & 33.32 & 7.05 & 7.05 & 17.88 & 11.64 & 4.07 & 13.50 & \textbf{32.99} & 13.78 & 6.94 & 18.89 & 15.00 & 5.42 & 15.50 & 18.43 & 9.49 & 4.14 & 5.34  \\
\algoName-PN++ & 41.25 &\textbf{44.92} &32.46 &21.64 &46.27 &19.52 &34.34 &21.04 &31.21 &31.91 &19.21 &36.14 &12.43 &25.32 &\textbf{21.37} &18.34 &21.61 &20.44  \\
\rowcolor[HTML]{EFEFEF} 
 \algoName \textbf{[ours]} & \textbf{49.26} & 41.36 &\textbf{36.21} &\textbf{28.64} &\textbf{58.31} &\textbf{19.68} &\textbf{38.91} & 21.98 &\textbf{38.10} &\textbf{35.54} &\textbf{21.03} &\textbf{41.61} &\textbf{16.19} &\textbf{29.07} & 21.09 &\textbf{24.68} &\textbf{24.13} &\textbf{23.30} \\
\midrule

\cellcolor[HTML]{ACCEAA}    \textbf{Weighted Modes Ratio \% $\uparrow$} & 
\cellcolor[HTML]{CBCEAA}    \includegraphics[width = 0.042\linewidth]{faucet.png} &
\cellcolor[HTML]{CBCEAA}    \includegraphics[width = 0.042\linewidth]{noun_Table_59987.png} &
\cellcolor[HTML]{CBCEAA}    \includegraphics[width = 0.042\linewidth]{noun_Cabinet_2881254.png} &
\cellcolor[HTML]{CBCEAA}    \includegraphics[width = 0.042\linewidth]{noun_Door_1549119.png} &
\cellcolor[HTML]{CBCEAA}    \includegraphics[width = 0.042\linewidth]{noun_window_3203560.png} &
\cellcolor[HTML]{CBCEAA}    \includegraphics[width = 0.042\linewidth]{noun_Fridge_1875643.png} &
\cellcolor[HTML]{ABCEBF}    \includegraphics[width = 0.042\linewidth]{avg.png} &

\cellcolor[HTML]{CBCECC}    \includegraphics[width = 0.042\linewidth]{faucet.png} &
\cellcolor[HTML]{CBCECC}    \includegraphics[width = 0.042\linewidth]{noun_Table_59987.png} &
\cellcolor[HTML]{CBCECC}    \includegraphics[width = 0.042\linewidth]{noun_Cabinet_2881254.png} &
\cellcolor[HTML]{CBCECC}    \includegraphics[width = 0.042\linewidth]{noun_Door_1549119.png} &
\cellcolor[HTML]{CBCECC}    \includegraphics[width = 0.042\linewidth]{noun_window_3203560.png} &
\cellcolor[HTML]{CBCECC}    \includegraphics[width = 0.042\linewidth]{noun_Fridge_1875643.png} &
\cellcolor[HTML]{ABCEBF}    \includegraphics[width = 0.042\linewidth]{avg.png} &
    
\cellcolor[HTML]{CBCEEE}    \includegraphics[width = 0.042\linewidth]{cooking-pot.png} &
\cellcolor[HTML]{CBCEEE}    \includegraphics[width = 0.042\linewidth]{kettle.png} &
\cellcolor[HTML]{CBCEEE}    \includegraphics[width = 0.042\linewidth]{switch.png} & 
\cellcolor[HTML]{ABCEBF}    \includegraphics[width = 0.042\linewidth]{avg.png} \\
  
\midrule
  Random Interaction & 5.61 &5.27 &7.62 &12.77 &15.61 &5.26 &8.69 &4.47 &3.12 &6.13 &9.73 &16.72 &3.92 &7.35 &13.32 &10.23 &5.87 &9.81\\
  \rowcolor[HTML]{EFEFEF} 
  Where2Act \cite{mo2021where2act} & 11.77 & 6.06 & 6.25 & 14.50 & 8.59 & 3.51 & 8.44 & 10.86 & 9.71 & 6.02 & 11.00 & 10.63 & 5.17 & 8.89 & 18.43 & 8.81 & 3.91 & 5.19 \\
\algoName-PN++ & 29.11 &25.52 &20.48 &12.81 &45.54 &17.48 &25.16 &14.28 &24.18 &26.66 &17.29 &25.92 &10.44 &19.79 &18.51 &15.76 &16.09 &16.79   \\
\rowcolor[HTML]{EFEFEF} 
 \algoName \textbf{[ours]} & \textbf{39.20} &\textbf{36.49} &\textbf{25.56} &\textbf{18.76} &\textbf{57.42} &\textbf{17.29} &\textbf{32.45} &\textbf{15.12} &\textbf{34.58} &\textbf{32.55} &\textbf{18.90} &\textbf{33.21} &\textbf{14.56} &\textbf{24.82} &\textbf{18.92} &\textbf{17.68} &\textbf{17.31} &\textbf{17.97}  \\
  
\midrule

\cellcolor[HTML]{ACCEAA}    \textbf{Weighted Normalized Entropy \% $\uparrow$} & 
\cellcolor[HTML]{CBCEAA}    \includegraphics[width = 0.042\linewidth]{faucet.png} &
\cellcolor[HTML]{CBCEAA}    \includegraphics[width = 0.042\linewidth]{noun_Table_59987.png} &
\cellcolor[HTML]{CBCEAA}    \includegraphics[width = 0.042\linewidth]{noun_Cabinet_2881254.png} &
\cellcolor[HTML]{CBCEAA}    \includegraphics[width = 0.042\linewidth]{noun_Door_1549119.png} &
\cellcolor[HTML]{CBCEAA}    \includegraphics[width = 0.042\linewidth]{noun_window_3203560.png} &
\cellcolor[HTML]{CBCEAA}    \includegraphics[width = 0.042\linewidth]{noun_Fridge_1875643.png} &
\cellcolor[HTML]{ABCEBF}    \includegraphics[width = 0.042\linewidth]{avg.png} &

\cellcolor[HTML]{CBCECC}    \includegraphics[width = 0.042\linewidth]{faucet.png} &
\cellcolor[HTML]{CBCECC}    \includegraphics[width = 0.042\linewidth]{noun_Table_59987.png} &
\cellcolor[HTML]{CBCECC}    \includegraphics[width = 0.042\linewidth]{noun_Cabinet_2881254.png} &
\cellcolor[HTML]{CBCECC}    \includegraphics[width = 0.042\linewidth]{noun_Door_1549119.png} &
\cellcolor[HTML]{CBCECC}    \includegraphics[width = 0.042\linewidth]{noun_window_3203560.png} &
\cellcolor[HTML]{CBCECC}    \includegraphics[width = 0.042\linewidth]{noun_Fridge_1875643.png} &
\cellcolor[HTML]{ABCEBF}    \includegraphics[width = 0.042\linewidth]{avg.png} &
    
\cellcolor[HTML]{CBCEEE}    \includegraphics[width = 0.042\linewidth]{cooking-pot.png} &
\cellcolor[HTML]{CBCEEE}    \includegraphics[width = 0.042\linewidth]{kettle.png} &
\cellcolor[HTML]{CBCEEE}    \includegraphics[width = 0.042\linewidth]{switch.png} & 
\cellcolor[HTML]{ABCEBF}    \includegraphics[width = 0.042\linewidth]{avg.png} \\
  
\midrule
  Random Interaction & 5.19 &4.45 &6.80 &10.49 &10.41 &4.18 &6.92 &7.09 &2.82 &4.89 &5.74 &11.30 &3.01 &5.81 &10.02 &7.41 &5.79 &7.74  \\
  \rowcolor[HTML]{EFEFEF} 
Where2Act \cite{mo2021where2act} & 12.12 & 5.08 & 5.41 & 9.03 & 5.15 & 3.23 & 6.67 & \textbf{15.62} & 8.31 & 5.93 & 6.75 & 5.30 & 4.23 & 7.68 & \textbf{17.84} & 7.60 & 3.91 & 4.89 \\
\algoName-PN++ & 24.60 &38.28 &28.48 &17.85 &32.66 &8.74 &25.10 &6.51 &13.02 &16.43 &7.22 &14.76 &6.00 &10.66 &15.78 &12.34 &12.40 &13.51 \\
\rowcolor[HTML]{EFEFEF} 
 \algoName \textbf{[ours]} & \textbf{34.79} &\textbf{36.49} &\textbf{35.64} &\textbf{25.48} &\textbf{41.76} &\textbf{9.31} &\textbf{30.58} & 7.14 &\textbf{19.38} &\textbf{24.15} &\textbf{7.77} &\textbf{19.27} &\textbf{8.16} &\textbf{14.31} &16.31 &\textbf{16.17} &\textbf{15.58} &\textbf{16.02} \\

  \bottomrule
  
  \end{tabular}
  }
\end{table*}

\subsection {Goal-conditional inference}
\label{subsec:goal-conditioned-method}

\algoName can be used for goal-conditional inference, by providing the desired goal observation $D_1$ as an extra input.
To do so, we replace the generative model with a deterministic one and fine-tune the system.
We treated $D_1$ as the conditional variable $g$ and re-format the training as followed, 
\begin{align}
    \tilde{\mathbb{P}}(a|s, g) = \underbrace{\mathbb{P}(a|o,z, g)}_{\text{action predictor}}~ \underbrace{\tilde{\mathbb{P}}(z|o, g)}_{\text{goal-conditional mode selector}}.
\end{align}
During the training, we keep the same action predictor and only fine-tune the mode selector $\mathbb{P}(z|o,g)$ to be goal-conditional.
This goal conditional mode selector produces the corresponding interaction mode latent $z$ for action predictor which in turn proposes action $a$ leading to goal $g$.

\section{Experiments}

Our experiments aim to evaluate the proposed method, \algoName, in terms of: 1) the ability to capture diverse interaction modes across varying object instances and categories, 2) its performance on unseen object states and instances from known categories, 3) its generalization on objects from unknown categories, and 4) the utility of the learned priors for goal-conditioned behaviors.

\subsection{Experimental setup}

Following~\cite{mo2021where2act}, we use articulated objects from the SAPIEN dataset~\cite{Xiang2020SAPIEN}. For training (seen categories), we pick nine categories: faucet, table, storage furniture, door, window, refrigerator, box, trashbin, and safe.  
For testing (unseen categories), we pick the 3 extra categories: kitchen pot, kettle, and switch.
For each category, we use 8 object instances with 4 different initial states for training and testing. 
We use IsaacGym~\cite{makoviychuk2021isaac} simulator to collect interaction data using a floating Franka parallel-jaw gripper. For collecting depth images, we vary the view angle of the camera in front of the object between  $\{ -45^{\circ}, -22.5^{ \circ}, 0^{\circ}, 22.5^{\circ}, 45^{\circ} \}$. To obtain the TSDF of the articulated object, we only use the given single depth image to compute the TSDF using ~\cite{zeng20163dmatch} during testing since we have found that using fewer cameras to reconstruct TSDF does not affect the testing results.

\paragraph{\textbf{Evaluation Metrics}}
To evaluate the multi-modal interaction modes, we use the following metrics to evaluate the prior distribution $\mathbb{P}(a|o)$:

\paragraph{1) \emph{sample-success-rate} ($ssr$)} measures the fraction of proposed interaction trials which are successful ~\cite{mo2021where2act}, 
\begin{align}
  ssr & = \frac{\text{\# successful proposed interaction}}{\text{\# total proposed interaction}}.
\end{align}
\paragraph{2) \emph{weighted modes ratio} ($\eta$)} measures the success rate weighted by the fraction of the interaction modes discovered,
\begin{align}
  \eta & = ssr \times \frac{\text{\# successful discovered mode}}{\text{\# total GT modes}}.
\end{align}
3) \paragraph{\emph{weighted normalized entropy} ($\bar{\mathcal{H}}$)} measures the success rate weighted by the entropy of the distribution.
\begin{align}
  \bar{\mathcal{H}} & =  ssr \times \frac{\mathcal{H}(\mathcal{M})}{\mathcal{H}_{max}},
\end{align}
where entropy $\mathcal{H}(\mathcal{M}) = -\sum_{m \in \mathcal{M}}p(m)\log p(m)$ is computed using $p(m)$, which is the percentage of sampled interactions leading to mode $m$. 
The maximum entropy $\mathcal{H}_{max}$ is computed under the condition of equally distributed proposed interaction modes.
Intuitively, for a more balanced prior distribution to sample rarer modes, $\bar{\mathcal{H}}$ should be higher since it covers possible interaction modes equally.

We use the ground-truth articulation state and part information to compute these metrics. We label an interaction successful when any object's DoF changes by at least $10\%$. Additionally, the weighted metrics are computed by verifying if interaction triggers a possible ground-truth interaction mode of the articulated object.



\begin{table*}[t] \LARGE
  \centering
  \caption{
  \small
  \textbf{Goal Conditional Evaluation}
  We evaluate our goal conditional model in terms of the success rate of reaching to the provided goal. We compare our model to Where2Act under 2 different evaluation tasks and bold the best number in each column section. 
  {\footnotesize Categories: 
  \hl{faucet, table, cabinet, door,  window, fridge},
  \hl{box, trash\_can, safe}
  }
  }
  \label{table:goal}
  \resizebox{1.0\linewidth}{!} 
  {
  \begin{tabular}{l|l||ccccccc|ccccccc|cccc}
  \toprule
\rowcolor[HTML]{FFDEB4}
\multicolumn{2}{l|}{\textbf{Test Set}} & \multicolumn{7}{c|}{\textbf{Unseen States of Training Objects}} & \multicolumn{7}{c|}{\textbf{Unseen instances of Training Categories}} & \multicolumn{4}{c}{\textbf{Unseen Categories}}\\ 
  \midrule

\cellcolor[HTML]{FFDEB4}
\textbf{Eval Task} &
\cellcolor[HTML]{ACCEAA}    \textbf{Sample Success Rate \% $\uparrow$} & 
\cellcolor[HTML]{CBCEAA}    \includegraphics[width = 0.042\linewidth]{faucet.png} &
\cellcolor[HTML]{CBCEAA}    \includegraphics[width = 0.042\linewidth]{noun_Table_59987.png} &
\cellcolor[HTML]{CBCEAA}    \includegraphics[width = 0.042\linewidth]{noun_Cabinet_2881254.png} &
\cellcolor[HTML]{CBCEAA}    \includegraphics[width = 0.042\linewidth]{noun_Door_1549119.png} &
\cellcolor[HTML]{CBCEAA}    \includegraphics[width = 0.042\linewidth]{noun_window_3203560.png} &
\cellcolor[HTML]{CBCEAA}    \includegraphics[width = 0.042\linewidth]{noun_Fridge_1875643.png} &
\cellcolor[HTML]{ABCEBF}    \includegraphics[width = 0.042\linewidth]{avg.png} &
    
\cellcolor[HTML]{CBCECC}    \includegraphics[width = 0.042\linewidth]{faucet.png} &
\cellcolor[HTML]{CBCECC}    \includegraphics[width = 0.042\linewidth]{noun_Table_59987.png} &
\cellcolor[HTML]{CBCECC}    \includegraphics[width = 0.042\linewidth]{noun_Cabinet_2881254.png} &
\cellcolor[HTML]{CBCECC}    \includegraphics[width = 0.042\linewidth]{noun_Door_1549119.png} &
\cellcolor[HTML]{CBCECC}    \includegraphics[width = 0.042\linewidth]{noun_window_3203560.png} &
\cellcolor[HTML]{CBCECC}    \includegraphics[width = 0.042\linewidth]{noun_Fridge_1875643.png} &
\cellcolor[HTML]{ABCEBF}    \includegraphics[width = 0.042\linewidth]{avg.png} &
    
\cellcolor[HTML]{CBCEEE}    \includegraphics[width = 0.042\linewidth]{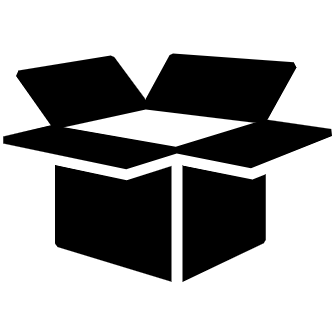} &
\cellcolor[HTML]{CBCEEE}    \includegraphics[width = 0.042\linewidth]{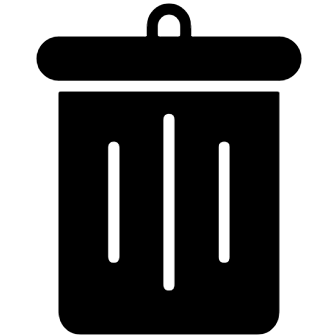} &
\cellcolor[HTML]{CBCEEE}    \includegraphics[width = 0.042\linewidth]{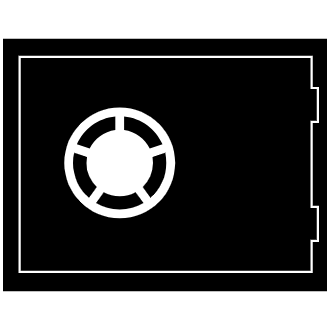} &
\cellcolor[HTML]{ABCEBF}    \includegraphics[width = 0.042\linewidth]{avg.png} \\
  
\midrule
  Dec-DoF& Where2Act-Push & \textbf{26.54} & 21.12 & 4.02 & 23.77 & 20.27 & 5.17 & 16.82 & 10.12 & 15.54 & 15.13 & \textbf{43.04} & 9.42 & 6.63 & 16.64 & 2.52 & \textbf{23.90} & \textbf{43.52} &23.31 \\
   \rowcolor[HTML]{EFEFEF} 
  (Common) & \algoName \textbf{[ours]} & 25.32 &\textbf{37.45} & \textbf{19.31} & \textbf{62.91} &\textbf{67.32 }& \textbf{61.23} &\textbf{45.59} &\textbf{20.31 }&\textbf{36.31 }&\textbf{18.24 }& 41.21 &\textbf{29.31 }&\textbf{31.42 }&\textbf{29.47 }&\textbf{15.17} & 22.50 & 32.51 &\textbf{23.39}  \\
  \hline
  Inc-DoF& Where2Act-Pull & 12.52 & 0.27 & 0.42 & 0.02 & 0.98 & 0.06 & 2.38 & 0.00 & 1.56 & 0.00 & 0.04 & 0.46 & 0.00 & 0.34 & 2.79 & 0.06 & \textbf{37.52} & 13.46\\
  \rowcolor[HTML]{EFEFEF} 
   (Rare Mode) & \algoName \textbf{[ours]} & \textbf{24.15} &\textbf{16.21} &\textbf{11.34} &\textbf{28.14} &\textbf{49.31} &\textbf{17.56} &\textbf{24.45} &\textbf{12.41} &\textbf{14.25} &\textbf{9.48} &\textbf{10.46} &\textbf{15.98} &\textbf{13.12} &\textbf{12.62} &\textbf{7.84} &\textbf{13.12} & 21.41 &\textbf{14.12}   \\

  \bottomrule
  
  \end{tabular}
  }
\end{table*}

\paragraph{\textbf{Baselines for comparisons and ablations}}
The proposed problem formulation in~\ref{sec:problem} is similar to skill-discovery in unsupervised RL (URL)~\cite{urlbench}. While the formulation is similar, URL is still to be shown effective in high-dimensional partially observed settings. In fact, we found that URL baselines including \cite{laskin2022cic,protorl,apt,diayn} performed poorer than a random policy and failed to express the complicated interaction modes. Another approach such as UMPNet~\cite{xu2022umpnet} looks at articulated object interaction in a goal-conditioned setting and relies on the groundtruth articulation state and part segmentation during training and inference. This is in stark contrast to our goal of learning priors without such privileged information.


Thus, we compare our approach to the following baselines:

\begin{enumerate}
  \item \textbf{Random policy}: uniformly samples interaction points from the articualted object's point cloud, orientation and moving direction. 
  \item \textbf{Where2Act} \cite{mo2021where2act}: computes priors per discretized action primitive. Since it evaluates interaction modes depending on separate primitives, we aggregrate the push and pull interactions and compute the average from these modes. Different from \cite{mo2021where2act}, we provide the whole object point cloud to the model instead of the segmented movable points as mentioned in the paper. 
  \item \textbf{\algoName-PN++}: is a version of ~\algoName that computes point features $v_{\textbf{p}}$ using PointNet++~\cite{qi2017pointnet} instead of ConvONet.
\end{enumerate}

\begin{figure}[!t]
    \centering
    \includegraphics[width=\linewidth]{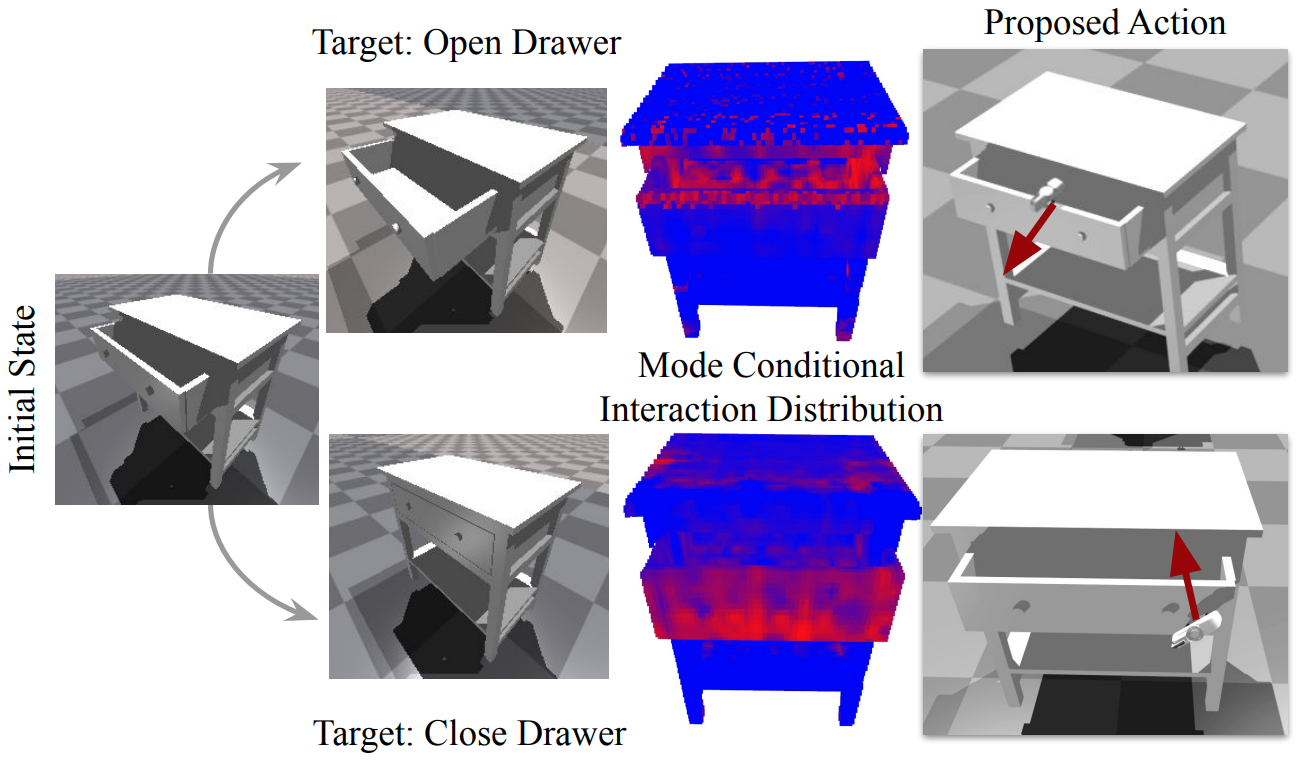}
    \caption{\textbf{Goal conditional manipulation results: } \algoName takes the and the target images to generate the mode conditional interaction distribution. Such distribution illustrates the valid part (colored in red) to manipulate. We sample the interaction point from pure point cloud based on the mode conditional point score and predict the action. The execution of the proposed action leads the initial state to the target state. }
    \label{fig:goal}
\end{figure}

\subsection{Goal conditional experiments}

As described in the goal-conditioned method, the  prior distribution can be used to infer goal-conditioned behaviors. 
After training the generative model in \algoName, we swap the input to the CVAE with a goal conditional input and fine-tune the trained model.
The goal conditional experiments are set with an extra input, goal observation $D_1$. 
We picked goal observation $D_1$ for each randomly selected object in a random pose and make sure that goal observation $D_1$ contains either the degrees-of-freedom increasing or decreasing.
We evaluate the goal-conditional success using sample-success-rate $ssr_{goal}$, defined as
\begin{align}
  ssr_{goal} & = \frac{\text{\# proposed interaction leads to goal $D_1$}}{\text{\# total proposed interaction}}.
\end{align}
We form the goals $D_1$ as increasing or decreasing object's DoF (from $D_0$), and report the average results per goal.
We consider Where2act \cite{mo2021where2act} as a baseline for goal-conditional inference since it explicitly defines the motion of pushing and pulling. 

\subsection{Discussion of Results}


\paragraph{\textbf{Interaction mode discovery}} We evaluate our self-supervised affordance mode discovery (i.e., non-goal-conditional inference) in Tab~\ref{table:self-supervised}. 
First, \algoName leads to a significantly more frequent successful interaction than the baselines, despite the fact that Where2Act uses access to ground-truth interaction modes during training.
Also, \algoName discovers more interaction modes than the baselines, based on the weighted normalized entropy.
Importantly, the competing method often trades interaction mode diversity for success rate, yet, \algoName leads at both metrics simultaneously.
We attribute our success to our task formulation, the advanced data collection strategy, and a stronger generative model.

In Fig.~\ref{fig:self-supervised}, we qualitatively show that \algoName discovers the correct interaction modes for different types of objects, even though it is trained without explicit knowledge of interaction modes. 
The distribution of interaction points provides a meaningful object segmentation into the movable part. 
Additionally, the interactions proposed by \algoName lead to a meaningful change in the states, resulting in different valid and useful interaction modes. 

\paragraph{\textbf{Goal-conditional inference}} To produce the goal-conditional interaction, we provide the model with an extra input $O_1$, the visual observation of the goal state.
As Table~\ref{table:goal} shows, our \algoName outperforms all the baselines on seen training categories. However, our results are slightly inferior to Where2Act~\cite{mo2021where2act} on the easier task $Dec-DoF$ (i.e., pushing), which is likely due to more training signals in Where2Act. Yet, on the rate events (i.e., pulling) our method still does best even on unseen objects. 
Figure \ref{fig:goal} visualized model outputs depending on the interaction goal. 
The goal-conditional interaction distribution shows how the modes of interaction change, depending on the given goal, which indicates that our method learned to map the goals to affordances.
The proposed action illustrates the correct way to grasp or touch the object and move in a reasonable direction. 
\section{Conclusions}
We propose a self-supervised method for discovering action affordances as modes of interaction for articulated objects, from purely vision observations.
\algoName generalizes across different modes of interactions and different categories of articulated objects.
Our method includes a novel adaptive data collection method, promoting interaction diversity, and a generative model to produce successful interactions with the objects, utilizing implicit object representations.
Our results show that our model generates interactions with a high success rate over a wide range of interaction modes, and can generalize to unseen objects and categories.

\section{Acknowledgement}
We would like to thank Kaichun Mo for the helpful discussions and Vector Institute for providing computation resources.
This work was supported in part by CIFAR AI Chair, NSERC Discovery Grant, and UofT XSeed Grant.

\clearpage
\renewcommand*{\bibfont}{\small}
\bibliographystyle{IEEEtran}
\bibliography{main}

\begin{thebibliography}{10}
\providecommand{\url}[1]{#1}
\csname url@samestyle\endcsname
\providecommand{\newblock}{\relax}
\providecommand{\bibinfo}[2]{#2}
\providecommand{\BIBentrySTDinterwordspacing}{\spaceskip=0pt\relax}
\providecommand{\BIBentryALTinterwordstretchfactor}{4}
\providecommand{\BIBentryALTinterwordspacing}{\spaceskip=\fontdimen2\font plus
\BIBentryALTinterwordstretchfactor\fontdimen3\font minus
  \fontdimen4\font\relax}
\providecommand{\BIBforeignlanguage}[2]{{%
\expandafter\ifx\csname l@#1\endcsname\relax
\typeout{** WARNING: IEEEtran.bst: No hyphenation pattern has been}%
\typeout{** loaded for the language `#1'. Using the pattern for}%
\typeout{** the default language instead.}%
\else
\language=\csname l@#1\endcsname
\fi
#2}}
\providecommand{\BIBdecl}{\relax}
\BIBdecl

\bibitem{abbatematteo2020kinematicmodels}
B.~Abbatematteo, S.~Tellex, and G.~Konidaris, ``Learning to generalize
  kinematic models to novel objects,'' in \emph{Proceedings of the Conference
  on Robot Learning}, ser. Proceedings of Machine Learning Research, vol.
  100.\hskip 1em plus 0.5em minus 0.4em\relax PMLR, 30 Oct--01 Nov 2020, pp.
  1289--1299.

\bibitem{klingbeil2010open}
E.~Klingbeil, A.~Saxena, and A.~Y. Ng, ``Learning to open new doors,'' in
  \emph{2010 IEEE/RSJ International Conference on Intelligent Robots and
  Systems}, 2010, pp. 2751--2757.

\bibitem{mittal2021articulated}
M.~Mittal, D.~Hoeller, F.~Farshidian, M.~Hutter, and A.~Garg, ``Articulated
  object interaction in unknown scenes with whole-body mobile manipulation,''
  2022.

\bibitem{gibson:hal-00692033}
J.~J. Gibson, ``{The theory of affordances},'' in \emph{{Perceiving, acting,
  and knowing: toward an ecological psychology}}.\hskip 1em plus 0.5em minus
  0.4em\relax {Hillsdale, N.J. : Lawrence Erlbaum Associates}, 1977, pp.
  pp.67--82.

\bibitem{gibson1986ecological}
J.~Gibson, \emph{The Ecological Approach to Visual Perception}, ser. Resources
  for ecological psychology.\hskip 1em plus 0.5em minus 0.4em\relax Lawrence
  Erlbaum Associates, 1986.

\bibitem{mo2021where2act}
K.~Mo, L.~Guibas, M.~Mukadam, A.~Gupta, and S.~Tulsiani, ``Where2act: From
  pixels to actions for articulated 3d objects,'' 2021.

\bibitem{wu2021vat}
R.~Wu, Y.~Zhao, K.~Mo, Z.~Guo, Y.~Wang, T.~Wu, Q.~Fan, X.~Chen, L.~Guibas, and
  H.~Dong, ``Vat-mart: Learning visual action trajectory proposals for
  manipulating 3d articulated objects,'' in \emph{ICLR}, 2022.

\bibitem{do2018}
T.-T. Do, A.~Nguyen, and I.~Reid, ``Affordancenet: An end-to-end deep learning
  approach for object affordance detection,'' in \emph{2018 IEEE International
  Conference on Robotics and Automation (ICRA)}, 2018, pp. 5882--5889.

\bibitem{xu2022umpnet}
Z.~Xu, H.~Zhanpeng, and S.~Song, ``Umpnet: Universal manipulation policy
  network for articulated objects,'' \emph{IEEE RA-L}, 2022.

\bibitem{KessensRSBG10}
C.~C. Kessens, J.~Rice, D.~Smith, S.~Biggs, and R.~Garcia, ``Utilizing
  compliance to manipulate doors with unmodeled constraints,'' in \emph{2010
  IEEE/RSJ International Conference on Intelligent Robots and Systems, October
  18-22, 2010, Taipei, Taiwan}.\hskip 1em plus 0.5em minus 0.4em\relax IEEE,
  2010, pp. 483--489.

\bibitem{opendoor}
A.~J. Schmid, N.~Gorges, D.~Goger, and H.~Worn, ``Opening a door with a
  humanoid robot using multi-sensory tactile feedback,'' in \emph{2008 IEEE
  International Conference on Robotics and Automation}, 2008, pp. 285--291.

\bibitem{servicerobots}
T.~Harada, A.~Tejero-de Pablos, S.~Quer, and F.~Savarese, ``Service robots: A
  unified framework for detecting, opening and navigating through doors,'' in
  \emph{Software Technologies}.\hskip 1em plus 0.5em minus 0.4em\relax Cham:
  Springer International Publishing, 2020, pp. 179--204.

\bibitem{generalized}
T.~Rühr, J.~Sturm, D.~Pangercic, M.~Beetz, and D.~Cremers, ``A generalized
  framework for opening doors and drawers in kitchen environments,'' in
  \emph{2012 IEEE International Conference on Robotics and Automation}, 2012,
  pp. 3852--3858.

\bibitem{kinematicefficient}
A.~Jain and S.~Niekum, ``Learning hybrid object kinematics for efficient
  hierarchical planning under uncertainty,'' 2019.

\bibitem{screwnet}
A.~Jain, R.~Lioutikov, C.~Chuck, and S.~Niekum, ``Screwnet:
  Category-independent articulation model estimation from depth images using
  screw theory,'' 2020.

\bibitem{kinematicmodels}
B.~Abbatematteo, S.~Tellex, and G.~Konidaris, ``Learning to generalize
  kinematic models to novel objects,'' in \emph{Proceedings of the Conference
  on Robot Learning}, ser. Proceedings of Machine Learning Research, vol.
  100.\hskip 1em plus 0.5em minus 0.4em\relax PMLR, 30 Oct--01 Nov 2020, pp.
  1289--1299.

\bibitem{learnopendoor}
E.~Klingbeil, A.~Saxena, and A.~Y. Ng, ``Learning to open new doors,'' in
  \emph{2010 IEEE/RSJ International Conference on Intelligent Robots and
  Systems}, 2010, pp. 2751--2757.

\bibitem{affordancecue}
M.~Stark, P.~Lies, M.~Zillich, J.~Wyatt, and B.~Schiele, ``Functional object
  class detection based on learned affordance cues,'' in \emph{Computer Vision
  Systems}, A.~Gasteratos, M.~Vincze, and J.~K. Tsotsos, Eds.\hskip 1em plus
  0.5em minus 0.4em\relax Berlin, Heidelberg: Springer Berlin Heidelberg, 2008,
  pp. 435--444.

\bibitem{affordancegeometric}
A.~Myers, C.~L. Teo, C.~Fermüller, and Y.~Aloimonos, ``Affordance detection of
  tool parts from geometric features,'' in \emph{2015 IEEE International
  Conference on Robotics and Automation (ICRA)}, 2015, pp. 1374--1381.

\bibitem{affordanceinfer}
\BIBentryALTinterwordspacing
H.~Kjellström, J.~Romero, and D.~Kragić, ``Visual object-action recognition:
  Inferring object affordances from human demonstration,'' \emph{Computer
  Vision and Image Understanding}, vol. 115, no.~1, pp. 81--90, 2011. [Online].
  Available:
  \url{https://www.sciencedirect.com/science/article/pii/S107731421000175X}
\BIBentrySTDinterwordspacing

\bibitem{affordancegrasp}
H.~O. Song, M.~Fritz, D.~Goehring, and T.~Darrell, ``Learning to detect visual
  grasp affordance,'' \emph{IEEE Transactions on Automation Science and
  Engineering}, vol.~13, no.~2, pp. 798--809, 2016.

\bibitem{affordanceseg}
\BIBentryALTinterwordspacing
T.~Nagarajan and K.~Grauman, ``Learning affordance landscapes for interaction
  exploration in 3d environments,'' 2020. [Online]. Available:
  \url{https://arxiv.org/abs/2008.09241}
\BIBentrySTDinterwordspacing

\bibitem{whatcanido}
\BIBentryALTinterwordspacing
K.~Khetarpal, Z.~Ahmed, G.~Comanici, D.~Abel, and D.~Precup, ``What can i do
  here? a theory of affordances in reinforcement learning,'' 2020. [Online].
  Available: \url{https://arxiv.org/abs/2006.15085}
\BIBentrySTDinterwordspacing

\bibitem{hierarchical2019manoury}
\BIBentryALTinterwordspacing
A.~Manoury, S.~M. Nguyen, and C.~Buche, ``Hierarchical affordance discovery
  using intrinsic motivation,'' in \emph{Proceedings of the 7th International
  Conference on Human-Agent Interaction}, ser. HAI '19.\hskip 1em plus 0.5em
  minus 0.4em\relax New York, NY, USA: Association for Computing Machinery,
  2019, p. 186–193. [Online]. Available:
  \url{https://doi.org/10.1145/3349537.3351898}
\BIBentrySTDinterwordspacing

\bibitem{manuelli2019kpam}
L.~Manuelli, W.~Gao, P.~Florence, and R.~Tedrake, ``kpam: Keypoint affordances
  for category-level robotic manipulation,'' 2019.

\bibitem{turpin2021gift}
D.~Turpin, L.~Wang, S.~Tsogkas, S.~Dickinson, and A.~Garg, ``Gift:
  Generalizable interaction-aware functional tool affordances without labels,''
  2021.

\bibitem{qin2019keto}
Z.~Qin, K.~Fang, Y.~Zhu, L.~Fei-Fei, and S.~Savarese, ``Keto: Learning keypoint
  representations for tool manipulation,'' 2019.

\bibitem{fang2018learning}
K.~Fang, Y.~Zhu, A.~Garg, A.~Kurenkov, V.~Mehta, L.~Fei-Fei, and S.~Savarese,
  ``Learning task-oriented grasping for tool manipulation from simulated
  self-supervision,'' 2018.

\bibitem{sundermeyer2021contactgraspnet}
M.~Sundermeyer, A.~Mousavian, R.~Triebel, and D.~Fox, ``Contact-graspnet:
  Efficient 6-dof grasp generation in cluttered scenes,'' 2021.

\bibitem{mahler2017dexnet}
J.~Mahler, J.~Liang, S.~Niyaz, M.~Laskey, R.~Doan, X.~Liu, J.~A. Ojea, and
  K.~Goldberg, ``Dex-net 2.0: Deep learning to plan robust grasps with
  synthetic point clouds and analytic grasp metrics,'' 2017.

\bibitem{khetarpal2021temporally}
K.~Khetarpal, Z.~Ahmed, G.~Comanici, and D.~Precup, ``Temporally abstract
  partial models,'' 2021.

\bibitem{khetarpal2020i}
K.~Khetarpal, Z.~Ahmed, G.~Comanici, D.~Abel, and D.~Precup, ``What can i do
  here? a theory of affordances in reinforcement learning,'' 2020.

\bibitem{danfei2021}
D.~Xu, A.~Mandlekar, R.~Martín-Martín, Y.~Zhu, S.~Savarese, and L.~Fei-Fei,
  ``Deep affordance foresight: Planning through what can be done in the
  future,'' in \emph{2021 IEEE International Conference on Robotics and
  Automation (ICRA)}, 2021, pp. 6206--6213.

\bibitem{myers2015}
A.~Myers, C.~L. Teo, C.~Fermüller, and Y.~Aloimonos, ``Affordance detection of
  tool parts from geometric features,'' in \emph{2015 IEEE International
  Conference on Robotics and Automation (ICRA)}, 2015, pp. 1374--1381.

\bibitem{mo2021o2oafford}
K.~Mo, Y.~Qin, F.~Xiang, H.~Su, and L.~Guibas, ``{O2O-Afford}: Annotation-free
  large-scale object-object affordance learning,'' in \emph{Conference on Robot
  Learning (CoRL)}, 2021.

\bibitem{jiang2021synergies}
Z.~Jiang, Y.~Zhu, M.~Svetlik, K.~Fang, and Y.~Zhu, ``Synergies between
  affordance and geometry: 6-dof grasp detection via implicit
  representations,'' 2021.

\bibitem{peng2020convolutional}
S.~Peng, M.~Niemeyer, L.~Mescheder, M.~Pollefeys, and A.~Geiger,
  ``Convolutional occupancy networks,'' 2020.

\bibitem{zeng20163dmatch}
A.~Zeng, S.~Song, M.~Nie{\ss}ner, M.~Fisher, J.~Xiao, and T.~Funkhouser,
  ``3dmatch: Learning local geometric descriptors from rgb-d reconstructions,''
  in \emph{CVPR}, 2017.

\bibitem{cvae}
K.~Sohn, H.~Lee, and X.~Yan, ``Learning structured output representation using
  deep conditional generative models,'' in \emph{Advances in Neural Information
  Processing Systems}, vol.~28, 2015.

\bibitem{Xiang2020SAPIEN}
F.~Xiang, Y.~Qin, K.~Mo, Y.~Xia, H.~Zhu, F.~Liu, M.~Liu, H.~Jiang, Y.~Yuan,
  H.~Wang, L.~Yi, A.~X. Chang, L.~J. Guibas, and H.~Su, ``{SAPIEN}: A simulated
  part-based interactive environment,'' in \emph{The IEEE Conference on
  Computer Vision and Pattern Recognition (CVPR)}, June 2020.

\bibitem{makoviychuk2021isaac}
V.~Makoviychuk, L.~Wawrzyniak, Y.~Guo, M.~Lu, K.~Storey, M.~Macklin,
  D.~Hoeller, N.~Rudin, A.~Allshire, A.~Handa \emph{et~al.}, ``Isaac gym: High
  performance gpu-based physics simulation for robot learning,'' \emph{arXiv
  preprint arXiv:2108.10470}, 2021.

\bibitem{urlbench}
M.~Laskin, D.~Yarats, H.~Liu, K.~Lee, A.~Zhan, K.~Lu, C.~Cang, L.~Pinto, and
  P.~Abbeel, ``Urlb: Unsupervised reinforcement learning benchmark,'' 2021.

\bibitem{laskin2022cic}
M.~Laskin, H.~Liu, X.~B. Peng, D.~Yarats, A.~Rajeswaran, and P.~Abbeel, ``Cic:
  Contrastive intrinsic control for unsupervised skill discovery,'' 2022.

\bibitem{protorl}
D.~Yarats, R.~Fergus, A.~Lazaric, and L.~Pinto, ``Reinforcement learning with
  prototypical representations,'' 2021.

\bibitem{apt}
H.~Liu and P.~Abbeel, ``Behavior from the void: Unsupervised active
  pre-training,'' 2021.

\bibitem{diayn}
B.~Eysenbach, A.~Gupta, J.~Ibarz, and S.~Levine, ``Diversity is all you need:
  Learning skills without a reward function,'' 2018.

\bibitem{qi2017pointnet}
C.~R. Qi, L.~Yi, H.~Su, and L.~J. Guibas, ``Pointnet++: Deep hierarchical
  feature learning on point sets in a metric space,'' 2017.

\end{thebibliography}

\newpage

\end{document}